\documentclass[journal]{IEEEtran}
\ifCLASSINFOpdf
\else
\fi
\usepackage{multirow}
\usepackage{algorithm}
\usepackage{algorithmic}
\usepackage{amsmath}
\usepackage{booktabs}
\usepackage{multirow}
\usepackage{amssymb}
\usepackage{pifont}
\usepackage{comment}
\usepackage[hyphens]{url}
\usepackage[breaklinks]{hyperref} 

\usepackage{graphicx}
\usepackage{booktabs}

\usepackage{array}
\newcolumntype{C}[1]{>{\centering\arraybackslash}p{#1}}

\usepackage{xcolor}

\usepackage{todonotes}

\makeatletter
\def\blfootnote{\xdef\@thefnmark{}\@footnotetext}
\makeatother

\begin{document}
%
\title{On Responsible Machine Learning Datasets with Fairness, Privacy, and Regulatory Norms}

\author{Surbhi~Mittal\textsuperscript{1}, Kartik~Thakral\textsuperscript{1}, Richa~Singh\textsuperscript{1}, Mayank~Vatsa\textsuperscript{1}*, \\ Tamar Glaser\textsuperscript{2}, Cristian~Canton~Ferrer\textsuperscript{2}, Tal~Hassner\textsuperscript{2} \protect\\
\textsuperscript{1}IIT Jodhpur, India,   \textsuperscript{2}Meta, USA
\thanks{
Email: \{mittal.5, thakral.1,  richa, mvatsa\}@iitj.ac.in,  \{tamarglaser, ccanton, thassner\}@meta.com.\protect \\ * corresponding author
}}

\onecolumn

\maketitle

\begin{abstract}
Artificial Intelligence (AI) has made its way into various scientific fields, providing astonishing improvements over existing algorithms for a wide variety of tasks. In recent years, there have been severe concerns over the trustworthiness of AI technologies. The scientific community has focused on the development of trustworthy AI algorithms. However, machine and deep learning algorithms, popular in the AI community today, depend heavily on the data used during their development. These learning algorithms identify patterns in the data, learning the behavioral objective. Any flaws in the data have the potential to translate directly into algorithms. In this study, we discuss the importance of \textit{Responsible Machine Learning Datasets} and propose a framework to evaluate the datasets through a \textit{responsible rubric}. While existing work focuses on the post-hoc evaluation of algorithms for their trustworthiness, we provide a framework that considers the data component separately to understand its role in the algorithm. We discuss responsible datasets through the lens of fairness, privacy, and regulatory compliance and provide recommendations for constructing future datasets. After surveying over 100 datasets, we use 60 datasets for analysis and demonstrate that none of these datasets is immune to issues of \textit{fairness}, \textit{privacy preservation}, and \textit{regulatory compliance}. We provide modifications to the ``datasheets for datasets" with important additions for improved dataset documentation. With governments around the world regularizing data protection laws, the method for the creation of datasets in the scientific community requires revision. We believe this study is timely and relevant in today's era of AI.
\end{abstract}

\section{Introduction}

With the proliferation of artificial intelligence (AI) and machine learning (ML) techniques, different nation-level projects and \textit{technology for good} programs are touching the lives of billions. These systems have provided incredibly accurate results ranging from face recognition of a million faces~\cite{facerec} to beating eight world champions at Bridge~\cite{AIbeatsBridge}. It has achieved superlative performance in comparison with experienced medical practitioners in identifying pneumonia and analyzing heart scans, among other medical problem domains \cite{CXRspottinhbyAI,HeartscanAI}. Recently, art generated by an AI algorithm won a fine arts competition~\cite{ArtbyAI}. While the systems are broadly accelerating the frontiers of smart living and smart governance, they have also shown to be riddled with problems such as bias in vision and language models, leakage of private information in social media channels, and adversarial attacks, including deepfakes. This problematic behavior has been affecting the trustworthiness of AI/ML systems. This has led to the design of the \textit{Principles of Responsible AI} which focus on designing systems that are safe, trustworthy, reliable, reasonable, privacy-preserving, and fair \cite{nitiaayog}.

\blfootnote{$\dagger$ All data was stored and experiments performed on IITJ servers by IITJ  faculty and students.}


\begin{figure}[t]
\centering
\includegraphics[width=0.5\linewidth]{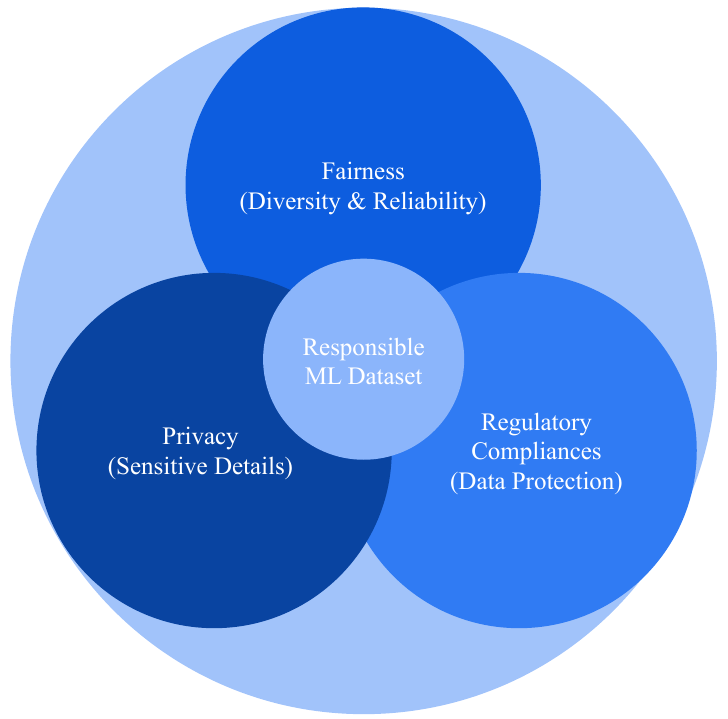}
\caption{We introduce the concept of Responsible Machine Learning Datasets and propose a quantitative rubric along with recommendations for future datasets.}
\label{fig:vizabstract}
\end{figure}

Among the different stages of an AI system development pipeline, data collection and annotation is one of the most important ingredients which can have a significant impact on the system. Current AI algorithms are deemed \textit{data-hungry} and tend to be extremely data-driven, and any irregularities in the datasets utilized during the development of these algorithms can directly impact the learning process. Several researchers have demonstrated that non-responsible use of datasets can lead to challenges such as fairness of the model and leakage of private information such as identity information or other sensitive attributes. Certain gender and race subgroups are shown to be under-represented in face-based image datasets~\cite{cao2018vggface2,yi2014learning} while some datasets contain objects specific to certain geographies or specific contexts~\cite{deng2009imagenet,rojasdollar}. Many algorithms have also been shown to suffer from spurious correlations in the dataset~\cite{geirhos2020shortcut,li2022whac,mehta2022you}. Similarly, concerns regarding the leakage of private information from popular datasets such as ImageNet have surfaced over recent years. In order to build responsible AI systems, it is therefore important to use datasets that are responsibly curated. We assert that \textit{Responsible Datasets leads to building Responsible AI Systems}.

Current research for understanding and evaluating trustworthiness focuses primarily on the performance of the models. However, by identifying these issues at the dataset level, we can lay the ground for creating better and \textit{responsible} datasets, and better AI. With the motivation to evaluate the reliability or trustworthiness of data, in this research, we present a framework to evaluate datasets via the proposed \textit{responsible rubric} across the axes of fairness, privacy, and regulatory compliance (refer to Figure \ref{fig:vizabstract}).
To the best of our knowledge, this is the first framework that quantitatively evaluates the trustability of the data used for training ML models. For defining dataset fairness, we consider the impact of three factors: diversity, inclusivity, and reliability of annotations. \textit{Inclusivity} considers whether different groups of people are present in the dataset across parameters of sex, skin tone, ethnic group, and age, and \textit{diversity} quantifies the distribution of these groups in the dataset.

For evaluating privacy preservation in datasets, we identify vulnerable annotations that can lead to the leakage of private information. Finally, we assess datasets for their degree of compliance with contemporary regulatory norms. Different governments around the world have approved various data privacy laws in the past few years. The popular General Data Protection Regulation (GDPR)~\cite{regulation2016regulation} requires the right to erasure based on its Article 17 (denoted as [Art. 17, GDPR]) and may require the consent of data subjects [Art. 12, GDPR] among other laws for data protection. Subjects providing data should have complete knowledge as to how their data will be used, and they should have the ability to revoke their consent. Further, if a dataset contains information such as images or potentially unethical data, there should be a mechanism to report such incidents. 
Through the axes of fairness, privacy and regulations, we demonstrate the applicability of the proposed framework by analyzing datasets from biometrics and healthcare domains, particularly face recognition and chest XRay datasets. After surveying over 100 datasets and discarding datasets unusable for this study
because of their small size or unavailability, we utilized a total of 60 datasets. Some of our key observations are as follows: 
\begin{itemize}
    \item Most of the existing datasets suffer on all three axes of \textit{fairness}, \textit{privacy} and \textit{regulatory compliance}, as per the proposed rubric. 
    \item Fairness is a major concern for most of the datasets we surveyed.
    \item Most existing datasets do not focus on regulatory compliances.
    \item Our analysis highlights that curating datasets from the web poses major risks to privacy preservation.
    \item  We also observe the \textit{fairness-privacy paradox} that exists in the development of datasets where the presence of sensitive attributes aids fairness evaluation but potentially leaks a subject’s private information.
\end{itemize}

Finally, we provide recommendations for constructing datasets. These recommendations can serve as an ethical sounding board for development of responsible datasets in the future.

\begin{table*}[]
\centering
\caption{\label{tab:regulations} Some of the laws surrounding data privacy around the world other than the GDPR~\cite{regulation2016regulation}.}
\begin{tabular}{|C{0.1\textwidth}|p{0.3\textwidth}|C{0.04\textwidth}|C{0.055\textwidth}|p{0.26\textwidth}|C{0.09\textwidth}|}
\hline
\textbf{Country(s)}       & \textbf{Title of Law}                                                                       & \textbf{From} & \textbf{Amended} & \textbf{Relevant Authority}                   & \textbf{Region}                    \\ \hline
Brazil                    & General Data Privacy Law (LGPD) \cite{brazil}                                              & 2018          & 2019             & Autoridade Nacional de Proteção de Dados (ANPD)      & Latin America                      \\
India                     & Information Technology Act 2000 \cite{indiait}                                             & 2000          & 2008             & Ministry of Law and Justice                          & Asia                               \\
                          & India's Personal Data Protection Bill (PDPB) 2019 \cite{indiapdpb}                         & 2019          & 2019             & Data Protection Authority of India                   &                                    \\
Israel                    & Protection of Privacy Law of 1981 \cite{israel}                                            & 1981          & 2020             & Privacy Protection Authority (PPA)                   & Middle East                        \\
Japan                     & Act on Protection of Personal Information \cite{japan}                                     & 2003          & 2020             & Personal Information Protection Commission           & Asia                               \\
Lithuania                 & Law on Legal Protection of Personal Data \cite{lithuania}                                  & 1996          & 2018             & State Data Inspectorate                              & Europe                             \\
New Zealand               & New Zealand's 1993 Privacy Act \cite{newzealand}                                           & 1993          & 2020             & Privacy Commissioner                                 & Australiasia                       \\
Nigeria                   & Nigeria Data Protection Regulation 2019 (NDPR) \cite{nigeria}                              & 2019          & 2019             & Nigerian Information Technology Development Agency   & Africa                             \\
South Africa              & Protection of Personal Information Act (POPIA) \cite{southafrica}                          & 2013          & 2013             & Information Regulator                                & Africa                             \\
Switzerland               & Data Protection Act \cite{switzerland}                                                     & 1992          & 2020             & Federal Data Protection and Information Commissioner & Europe                             \\
Thailand                  & Personal Data Protection Act (PDPA) 2019 \cite{thailand}                                   & 1997          & 2019             & Personal Data Protection Committee                   & Asia                               \\
Turkey                    & Law 6698 on Personal Data Protection (LPDP) \cite{turkey}                                  & 2016          & 2016             & Data Protection Authority                            & Europe                             \\
USA: California & California Privacy Rights Act of 2020, amending California Consumer Privacy Act \cite{usc} & 2020          & 2020             & California Privacy Protection Agency (CPPA)          & North America                      \\
USA: Illinois   & Biometric Information Privacy Act (BIPA) \cite{bipa}                                       & 2008          & -                & U.S. Court of Appeals for the Ninth Circuit    & \multicolumn{1}{l|}{North America} \\
USA: Texas      & Texas Consumer Privacy Act (TXCPA) \cite{texas}                                            & 2019          & -                & Texas House of Representatives                       & \multicolumn{1}{l|}{North America} \\
                          & Texas Capture or Use of Biometric Identifier Act (CUBI) \cite{texascubi}                   & 2009          & 2017             & Texas Attorney General                             & \multicolumn{1}{l|}{}              \\
USA: Washington & Washington Biometric Privacy Protection Act (House Bill 1493)  \cite{washington}           & 2017          & -                & Washington State Attorney General             & \multicolumn{1}{l|}{North America} \\ \hline
\end{tabular}
\end{table*}


\section{Related Work}
In recent years, there has been an increasing focus on datasets being used in ML and deep learning. Specifically, such concerns are heightened when datasets pertain to the collection of sensitive data such as biometrics and medical data. According to a recent study, the importance of dataset quality being used in AI/ML is constantly undermined, and the data collection work is often undervalued in the community~\cite{sambasivan2021everyone}.

Researchers have addressed the need for data-centric AI as well as the impact of regulations and policies on trustworthy AI~\cite{liang2022advances}. Heger et al.~\cite{heger2022understanding} conducted interviews with ML practitioners and discovered the need to emphasize the relationship between data documentation and responsible AI. Scheuerman et al. identified the patterns followed in the collection process of computer vision datasets based on 1000$+$ publications and also emphasized the importance of proper dataset documentation~\cite{scheuerman2021datasets}.

There has been discussion around the collection of socio-cultural data where researchers have highlighted the need to design institutional frameworks and procedures inspired by archival data~\cite{jo2020lessons}. Some of the essential considerations are consent, inclusivity, power, transparency, ethics, and privacy. Similarly, focusing on the entire dataset development pipeline, Peng et al.~\cite{peng2021mitigating} studied nearly 1000 papers citing problematic datasets such as Labeled Faces in the Wild (LFW), MS-Celeb-1M~\textit{(decommissioned)}, and DukeMTMC~\textit{(decommissioned)}. The authors provide recommendations for dataset creators as well as conference program committees to encourage more ethical creation of datasets. 

In 2021, Gebru et al.~\cite{gebru2021datasheets} proposed a comprehensive `datasheet' detailing information about the dataset accompanying its release. The datasheets are designed to raise transparency and accountability in the datasets. On similar lines, Hutchinson et al. introduced a framework that helps build accountability and transparency in the data development process~\cite{hutchinson2021towards}. By segregating data development process into various stages, the authors described the roles played by individuals such as requirements owner, stakeholder, and reviewer at each stage. Palluda et al.~\cite{paullada2021data} promoted the usage of quantitative as well as qualitative measures for the effective development of datasets. They showcase how representational harms and spurious correlations present in the datasets can lead to unfair decisions.

\noindent \textbf{Fairness in Datasets:} In order to build fairer AI, researchers have studied bias in various settings~\cite{tommasi2017deeper, yangfei2020towards, de2019does, singh2022anatomizing}. A recent report by NIST for identifying and managing bias in AI has cited the reliance on large-scale datasets as the leading cause of using unsuitable datasets for training and evaluation~\cite{schwartz2022towards}. The report discusses various challenges and factors associated with datasets in modern AI such as lack of representation, and statistical and socio-technical factors. Towards better representation in AI, Kamikubo et al. analyzed 190 accessibility datasets. Their analysis revealed harmful trends such as lack of older adults for autism, developmental and learning communities. To this end, they suggested more meaningful interactions with data contributors~\cite{kamikubo2022data}. Similarly, in the domain of NLP, some researchers have proposed the use of data statements that can help understand the intent and, specifically, the biases of the data~\cite{bender2018data}. The data statement emphasizes the inclusion of information such as the curation rationale and annotator demographic. Some researchers have proposed toolkits for evaluation of bias in the dataset which include object-based, person-based, and geography-based analyses through annotations~\cite{wang2022revise}.

\noindent \textbf{Privacy Leakage in Datasets:}
In this work, we refer to the term \textit{privacy leakage} as ``the unintended or unauthorized/accidental exposure of sensitive or protected personal data/information, which may compromise an individual's identity." The concerns of privacy leakage surrounding datasets in deep learning have grown over the past few years. In Birhane et al.~\cite{birhane2021large}, the authors discuss the issues of consent and privacy breaches in the context of large-scale vision datasets such as ImageNet. They highlight the harms associated with poor dataset curation practices and propose mandatory institutional reviews. 

In the field of security and privacy, researchers have attempted to preserve privacy by adopting different concepts, such as $k$-anonymity \cite{samarati1998protecting} and differential privacy \cite{dwork2008differential}. To quantify privacy leakage, researchers have proposed various metrics, such as $l$-diversity \cite{machanavajjhala2007diversity}, $k$-anonymity \cite{samarati1998protecting}, $t$-closeness \cite{li2006t}, and m-invariance \cite{xiao2007m}, among others. A detailed list of these metrics has been provided by Wagner et al.\cite{wagner2018technical}. These metrics are designed to capture the extent of privacy leakage from the perspective of an adversary with knowledge \cite{li2006t}, equivalent representation of sensitive attributes \cite{machanavajjhala2007diversity}, protection from homogeneity attacks \cite{samarati1998protecting}, and other reasons. 

After the introduction of $k$-anonymity \cite{sweeney2002k}, several researchers have developed techniques for facial privacy preservation. Zang et al. \cite{zhang2018privacy} devised a function to add random noise to existing data samples to synthesize new samples. This was aimed at masking sensitive information in the dataset while preserving the performance of the model. Chhabra et al. \cite{chhabra2018anonymizing} proposed an algorithm that provides the control to the user to anonymize k-facial attributes while preserving other attributes and identity information. Li et al. \cite{li2019anonymousnet} proposed a technique to anonymize identity and attribute while maintaining the data utility. The authors also performed quantification of privacy preservation through $k$-anonymity. 

In other works, researchers have used Mechanical Turk participants to identify privacy-sensitive information in images to automate the privacy attribute selection/identification task for obfuscation~\cite{li2018human}. Gervais et al.~\cite{gervais2016quantifying} propose a framework to infer the location by analysing the consumer purchase histories. Further, experiments have demonstrated the benefit of teaching algorithms to predict the presence and purpose of private information. Orekondy et al.~\cite{orekondy2017towards} propose an algorithm to predict a leakage risk score for the input image using their proposed VISPR dataset. Orekondy et al.~\cite{orekondy2018connecting} have also proposed a redaction by segmentation approach to aid users in selectively sanitizing images of private content. The proposed approach segments out the privacy attributes in images and provides privacy vs data-utility evaluation analysis. Gurari et al.~\cite{gurari2019vizwiz} propose a dataset and study visual privacy issues faced by people who are blind and are trying to learn about their physical surroundings. \\

\noindent \textbf{Regulatory Compliance in Datasets:} With the increasingly high emphasis on data protection, various countries around the world have put legislation in place for data security and privacy. According to a report, 157 countries in the world had instated data privacy laws by mid-March 2022~\cite{unctad, greenleaf2021global, greenleaf2022now}. Most of these laws are influenced by GDPR~\cite{regulation2016regulation} but contain certain variations.
The GDPR may prohibit the processing of biometric data unless explicit consent from the subjects is not provided. By providing the \textit{right to be forgotten}, the GDPR puts the subject in charge of their data. Different studies have been conducted to understand the impact of GDPR on artificial intelligence~\cite{forti2021deployment, goldsteen2021data}. 

In Table~\ref{tab:regulations}, we summarize data privacy laws for some of the countries around the world. There are other laws specific to certain kinds of data, such as the Health Insurance Portability and Accountability Act (HIPAA)~\cite{act1996health} for medical health in the US and the Biometric Information Privacy Act (BIPA)~\cite{bipa} protecting biometric information in the state of Illinois, US. Other US states are actively working towards enforcing their data privacy laws~\cite{newUSlaws}. The privacy implications of some of these acts have also received significant attention in the research community~\cite{nosowsky2006health}. Notably, the European Commission released \textit{Ethics Guidelines for Trustworthy AI}, which discusses the framework, foundations, and possible assessments for trustworthy AI~\cite{ethicsai}, as well as is in the process of amending and debating the Artificial Intelligence Act~\cite{euaiact} to address risks associated with AI applications. Recent work discusses the impact of the Artificial Intelligence Act on facial processing applications~\cite{hupont2022landscape}.

While different works identify different problematic aspects of dataset collection, very few works have looked at the factors of fairness, privacy, and regulatory compliance in datasets holistically. In this work, we provide quantitative as well as qualitative insight across the three factors and how data collection in AI needs to turn towards better and more responsible datasets.



\begin{table}[]
\centering
\caption{The different demographic subgroups considered for fairness quantification.}
\begin{tabular}{|p{0.08\textwidth}|p{0.35\textwidth}|}
\hline
\textbf{Demography} & \textbf{Subgroups}                                                                                                      \\ \hline
Sex                 & Male, Female, Other                                                                                                     \\ 
\multirow{3}{*}{Ethnicity}           & White/Caucasian, Black/African, Southeast Asian, East Asian, Indian, Hispanic/Latino, Middle Eastern, Mixed-race, Other \\ 
Skin tone           & Type I-VI using Fitzpatrick scale                                                                                       \\ 
\multirow{2}{*}{Age} & 0-3, 4-7, 8-15, 16-20, 21-30, 31-40, 41-50, 51-60, 61-70, 71-100                                                        \\ \hline
\end{tabular}
\label{tab:fairnessgroups}
\end{table}

\section{Methods}
In this section, we describe the methodology adopted for designing the framework for \textit{Responsible Datasets}. We quantify datasets across the axes of fairness, privacy, and regulatory compliance. The concerns regarding these factors may vary from domain to domain. For example, fairness in a face image dataset may differ from those in an object or egocentric dataset. The quantification in this section is based on datasets centered around individuals and specifically, face-based datasets.

\subsection{Quantifying Dataset Fairness} In deep learning, fairness concerns have been raised for datasets in multiple domains in different contexts~\cite{cao2018vggface2,rojasdollar}. In face-based image datasets, some sex and race subgroups may be under-represented~\cite{cao2018vggface2,yi2014learning}. In object-based datasets, datasets may contain objects specific to certain geographies or in specific contexts~\cite{rojasdollar,deng2009imagenet}. Similarly, text-based and multi-modal datasets may suffer from spurious correlations in datasets leading to bias in the performance of trained models~\cite{geirhos2020shortcut}.

\begin{figure*}[]
\centering
\includegraphics[width = 0.97\textwidth]{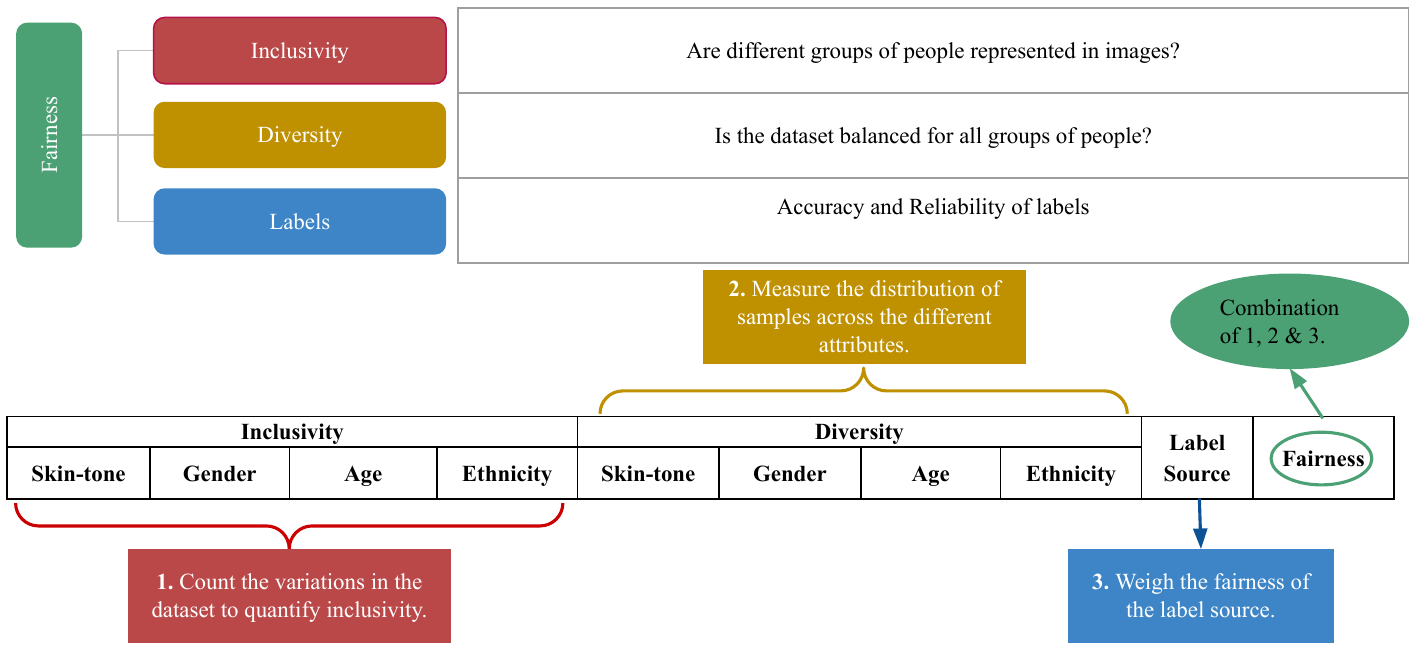}
\caption{(Top) The three aspects involved in fairness quantification- Inclusivity, Diversity, and Labels, and the questions they answer. (Bottom) The formulation employed for the calculation of the fairness score.}
\label{fig:fairformula}
\end{figure*}

In this work, we consider the impact of three factors for quantification of dataset fairness- diversity, inclusivity, and labels (See Fig.~\ref{fig:fairformula}). In the context of face-based datasets, \textit{inclusivity} quantifies whether different groups of people are present in the dataset across parameters of sex, skin tone, ethnic group, and age. \textit{Diversity} quantifies the distribution of these groups in the dataset, with an assumption that a balanced dataset is the most fair. While a balanced dataset does not guarantee equal performance, existing work has shown improved fairness with the use of balanced datasets \cite{wang2021meta, ramaswamy2020fair}. We note that such a dataset may not be ideal in many cases, but it acts as a simplifying assumption for the proposed formulation. Finally, we consider the reliability of the \textit{labels} depending on whether they have been self-reported by the subjects in the dataset or are annotated based on apparent characteristics. 

We consider four demographic groups- \textit{sex}, \textit{skin tone}, \textit{ethnicity}, and \textit{age}. The different subgroups considered for these demographics are specified in Table~\ref{tab:fairnessgroups}. We utilize the information regarding the biological sex of an individual while leaving room for error in the class \textit{Other}. For ethnic subgroups, we take inspiration from the FairFace dataset~\cite{karkkainen2021fairface} with the addition of the mixed-race as a separate category. Ethnicity subgroups around the world tend to be extremely variable. We have adopted the maximum ethnic subgroups as represented in the literature by the FairFace dataset. The age subgroup classification is based on the categories in the AgeDB dataset~\cite{moschoglou2017agedb}. The age annotations were binned as per AgeDB categorization for datasets that provided continuous age values.

The proposed formulation for quantification of fairness in a dataset is dependent on the annotations available in a dataset (See Fig.~\ref{fig:fairformula}). Let \textbf{D} = \{sex, skintone, ethnicity, age\} denote the complete set of demographics considered for evaluation of any dataset, and \textbf{S} denote the corresponding subgroups in each demographic (Refer Table \ref{tab:fairnessgroups} for subgroups considered for each demographic). Then, $\textbf{D}_1$ = sex, and $\textbf{S}_1$= \{male, female, other\}. For a given dataset, \textbf{d} denotes the set of demographics annotated in the dataset, and \textbf{s} denotes the subgroups corresponding to those demographics. For example, for the AgeDB dataset \cite{moschoglou2017agedb}, \textbf{d} = \{sex, age\}, and $\textbf{s}_i$ = \{male, female\} for $ith$ demographic in \textbf{d}, and  $\textbf{s}_{ij}$ = male for $jth$ subgroup of $ith$ demographic $(i=1, j=1)$. Then, the \textbf{inclusivity} $r_i$ for each demography is defined as the ratio of demographic subgroups present in the dataset and the pre-defined demographic subgroups in $\textbf{S}_i$. This is quantified as-
\begin{equation}
     r_i = 	|s_i|/|S_i|
\end{equation}

The \textbf{diversity} $v_i$ is calculated using Shannon's diversity index~\cite{shannon1948mathematical} to capture the distribution of different subgroups for a given demography $d_i$ as follows, 
\begin{gather}
    p_{ij} = num(s_{ij})/\sum_j{num(s_{ij})}  \\
    v_i = -  \frac{1}{ln(|s_i|)} \sum_j{p_{ij} * ln(p_{ij})},
\end{gather}

where $num(\textbf{s}_{ij})$ denotes the number of samples for the $jth$ subgroup of the $ith$ demographic in the dataset. In certain cases where the number of samples is not available, we consider $num$ to denote the number of subjects in the dataset. Fairness across each of the demographics is measured between 0 to 1. For example, if a dataset contains images corresponding to each of the six skin tones, it will have an inclusivity score of 6/6 = 1, and if the number of samples is balanced across each of the subgroups of skin tone, the diversity score will also be 1. When combined, that will provide an overall score of 1*1 = 1. By multiplying the inclusivity and diversity scores, we are providing information about the presence as well as distribution of samples corresponding to a given demographic. These scores are added for the four demographic attributes.

The \textbf{label score} $l$, is then calculated based on whether the labels or annotations for the dataset are self-reported, classifier-generated, or apparent. \textit{Self-reported labels} indicate that the subjects provided their demographic information as a part of the data collection process. \textit{Classifier-generated labels} imply that the demographic labels were obtained through an automated process of classification. Finally, \textit{apparent labels} indicate that the annotations were done by an external annotator after observing the images (for example, through Amazon Mechanical Turk). Based on the type of annotations, a \textit{label score} is assigned for the dataset between 0 to 1, with self-reported labels assigned a value of 1, classifier-generated labels assigned 0.67, and apparent labels assigned a value of 0.33. machine-generated labels, sometimes referred to as {\em pseudo labels}, are assigned a higher score since they have been shown to be more reliable than human-annotated labels for use in deep learning-based applications \cite{lee2013pseudo, tuan2017regressing, chang2019deep}. We acknowledge that there may be different perceptions of reliability in annotation based on the task and nature of the data. However, the labels generated by a trained classifier are consistent while that may not be true for human annotators. Based on this rationale, we assign a higher score to classifier-generated labels. 

In cases where the labels are collected using more than one of the three categories, an average of the corresponding categories' scores is taken. For medical datasets, a score of 1 is provided if a medical professional provides/validates the annotations, else a score of 0 is provided. The label score is provided for the entire dataset as per the current formulation. To calculate the \textbf{fairness score}, $F$, for the dataset, the factors of inclusivity, diversity, and labels are combined as follows,
\begin{equation} \label{eq:fairnessquant}
    F = \sum_i{(r_i * v_i)} + l
\end{equation}
The fairness score is designed such that a higher value indicates a fairer dataset while a lower value indicates a less fair dataset.

\subsection{Quantifying Dataset Privacy} 
\begin{figure}[t]
\centering
\includegraphics[width=0.43\textwidth]{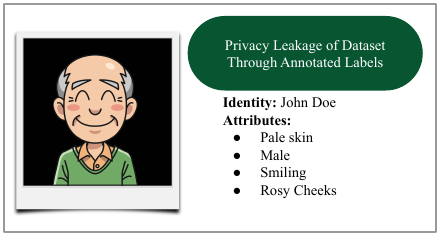}
\caption{Privacy leakage through the information available in datasets. The sample is representative of information present in datasets such as the LFW dataset \cite{huang2008labeled, liu2015deep}.}
\label{fig:privacy}
\end{figure}

Deductions and attacks can be carried out using the annotated labels in publicly available supervised datasets (See Figure \ref{fig:privacy}). Annotated attributes in a face dataset, for example, can be used for face profiling~\cite{findface, facepp, socialmapper, gnanasekar2019face}. In contrast, vehicle registration numbers from location datasets can be used to make malicious deductions and track someone down~\cite{orekondy2017towards,5958033}. As a result, the more annotations there are in the dataset, the more privacy is potentially leaked. The extent of privacy leakage can be summarised in terms of the quantity of information leaked and the extent to which private information is exposed in the annotated labels. 

\noindent In this work, for quantification of privacy leakage in the publicly available datasets, we identify vulnerable label annotations that can lead to the potential leakage of private information and devise a mathematical formulation. This formulation employs the dataset's annotated labels to quantify the potential privacy leakage. We identify six label annotations that are widely available in the datasets and lead to leakage of privacy in datasets: \textit{name identification}, \textit{sensitive and protected attributes}, \textit{accessories}, \textit{critical objects}, \textit{location inference}, and \textit{medical condition}. 
Let the set $A$ constitute these identified attributes, i.e.:

\begin{equation}
    A = \{A_N, A_{SP}, A_{AC}, A_C, A_L, A_M\}
\end{equation}

The attributes used in defining $A$ are as follows:
\begin{itemize}
    \item \textit{Name Identification Information $A_N$:} This attribute refers to the name of each individual annotated in the dataset. This annotation potentially leads to the highest level of privacy leakage.
    \item \textit{Sensitive and Protected Attribute Information $A_{SP}$:} These attributes refer to information regarding gender, sexual orientation, race, past records, etc., corresponding to an individual.
    \item \textit{Accessory Information $A_{AC}$:} This attribute denotes the presence of accessories such as hats and sunglasses in a face image as well as other attributes such as five 'o'clock shadow. 
    \item \textit{Critical Objects $A_C$:} This attribute denotes the presence of objects revealing the identity of a person, such as credit cards or signatures.
    \item \textit{Location Information $A_L$:} This attribute denotes the presence of information in the image that can potentially disclose a person's location, such as geographical coordinates or popular landmarks in the image background.
    \item \textit{Medical Condition Information $A_M$:} This attribute denotes the presence of any information regarding the medical condition of an individual in the dataset.
\end{itemize}

For a dataset, we manually check for the presence of the annotations from the aforementioned list to estimate its privacy leakage, and a point is awarded for each attribute annotation. The privacy leakage score ($PL$) is then calculated as the sum of all the present attributes as described below:

\begin{equation}
    PL =\sum_{i=1}^{6} A_i.
\end{equation}

Finally, the \textbf{privacy preservation score}, $P$, for a given dataset is estimated as,
\begin{equation}
    P = (|A| - PL).
\end{equation}

$P$ indicates the amount of information being preserved in a dataset and represents its dependability for public use. We note that while the presence of these annotations constitutes privacy leakage in our formulation, it aids the computation of the fairness score described in the previous section. We discuss this fairness-privacy paradox in detail later in the text.

\subsection{Quantifying Regulatory Compliance} Different countries around the world have approved various data privacy laws in the past few years. One of the most widely accepted documents covering data privacy laws is the one applicable in European countries known as the GDPR~\cite{regulation2016regulation}. The following laws can be applied to deep learning methodologies and/or datasets,
\begin{itemize}
    \item Right to be forgotten (right to erasure) [Art. 17]
    \item Consent of data subjects [Art. 12]
    \item Right to object/restriction to the processing of their data [Art. 5, 6, 9, 18, 19]
    \item Right to rectification [Art. 16]
    \item Right of access by the data subject [Art. 15]
    \item Right to object and automated individual decision-making [Art. 21 and 22] [Recital 71]
    \item Right to lodge complaint [Art. 77]
    \item Right to effective judicial remedy [Art. 78,79]
\end{itemize}
\noindent Some of these laws restrict the use of users' personal data unless their consent is available for that particular application [Art. 5, 6, 9, 18, 19]. Other laws and conditions specified in the GDPR include,
\begin{itemize}
    \item Right to data portability [Art. 20]
    \item Security of personal data [Art. 32, 33, 34]
    \item Conditions for consent [Art. 7]
    \item Conditions for protection of children's personal data [Art. 8]
    \item Requirements for regular data protection impact assessments (DPIA) [Art. 35]
    \item Cryptographic protection of sensitive data  
    \item Breach notification requirements. [Art. 33]
\end{itemize}

Apart from the laws specified above, datasets in deep learning can benefit from existing mechanisms for institutional approval (such as IRB) and newer requirements set by popular conferences such as ethics and impact statement for datasets~\cite{EthicsCVPR}. In this paper, the \textbf{regulatory compliance score}, $R$,  in the dataset is quantified based on three factors- institutional approval (yes/no: the numerical value of 1/0), the subject's consent to the data collection (yes/no: the numerical value of 1/0), and the facility for expungement/correction of the subject's data from the dataset (yes/no: the numerical value of 1/0). If a dataset satisfies all three criteria, a compliance score of $3$ is provided. While the absence of a data subject's consent may not necessarily breach regulatory norms, for lack of a more subtle evaluation, we utilize \textit{subject consent} in the dataset as one of the factors for compliance. For example, the privacy rule in HIPAA compliance does not restrict the distribution of de-identified health data. The different factors for compliance are manually validated via information present in the published paper, webpage, and/or GitHub page for the dataset. Unless the information is explicitly specified in the aforementioned resources, it is assumed to be absent in which case we assign a value of zero. 

\begin{figure*}[!]
\centering
\includegraphics[width=\textwidth]{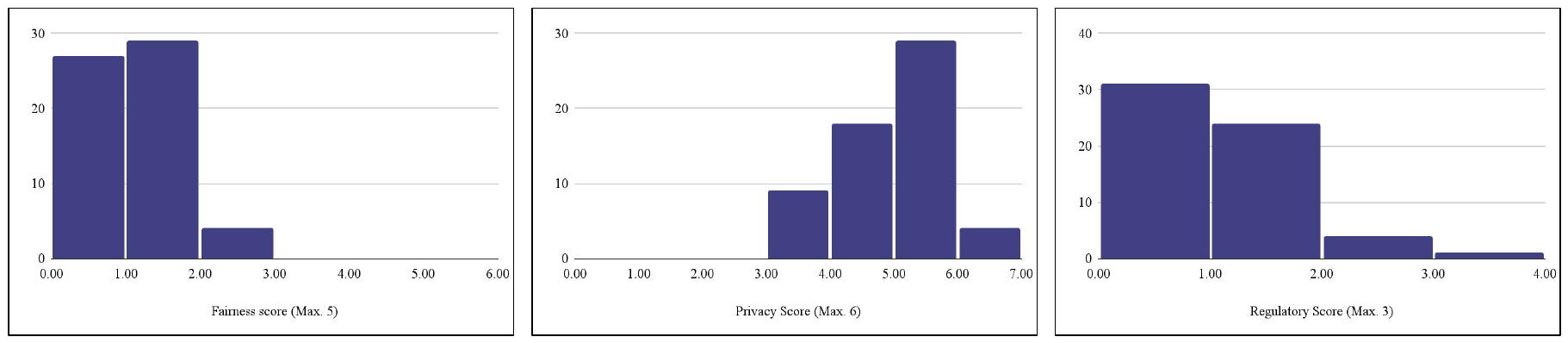}
\caption{The summary of fairness, privacy, and regulatory compliance scores through histogram visualization for the datasets we surveyed. (Left) The maximum value of the fairness score that can be obtained is 5, but it is observed that the fairness scores do not exceed a value of 3. (Middle) While most datasets in our study preserve privacy in terms of not leaking location or medical information, very few provide perfect privacy preservation. (Right) Most datasets comply with no regulatory norm or only one. We can observe from this plot that most datasets provide a low fairness score and perform poorly on the regulatory compliance metric.}
\label{fig:barplotsummary}
\end{figure*}

\begin{figure*}[!t]
\centering
\includegraphics[width=0.8\textwidth]{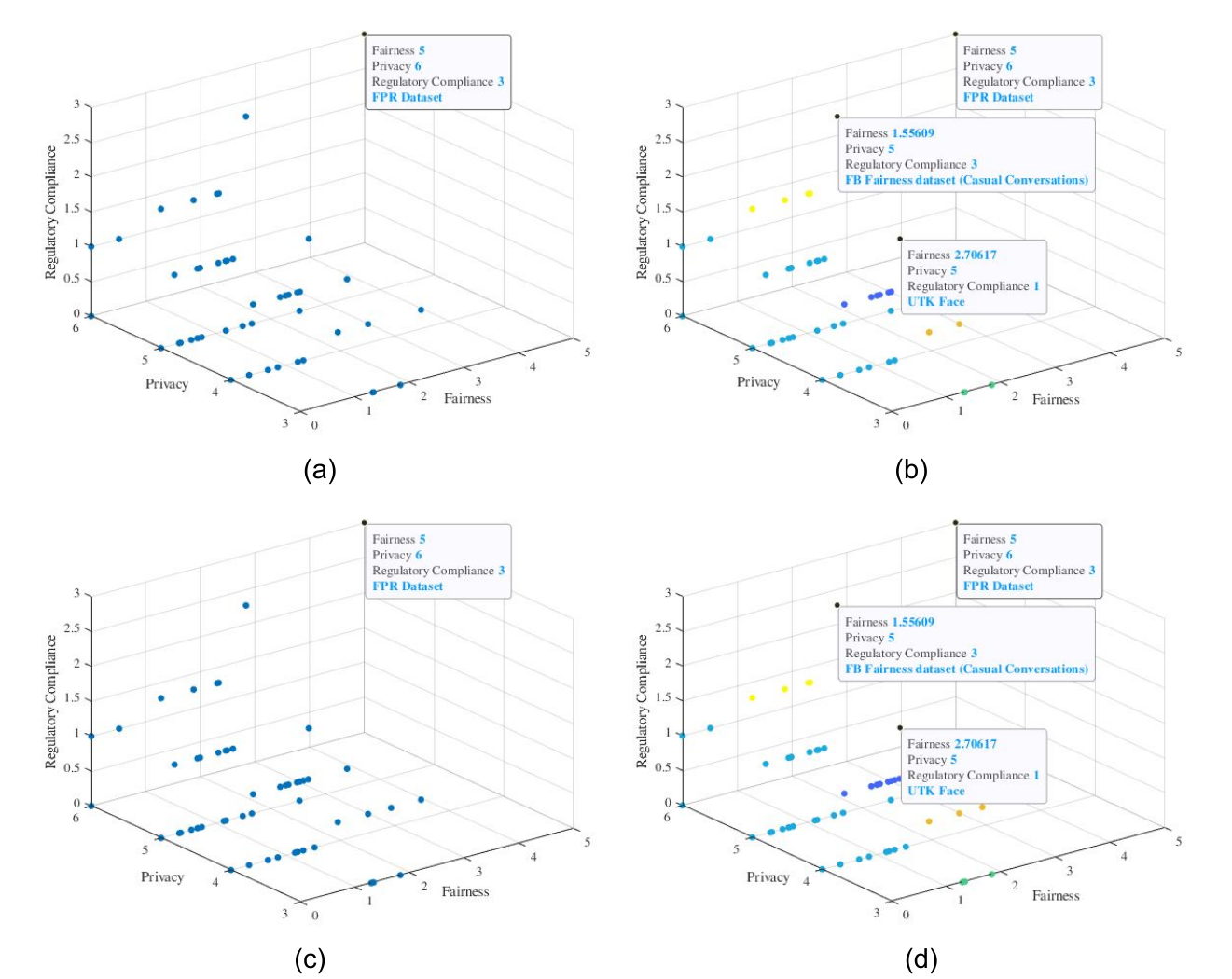}
\caption{Cluster analysis based on the 3-tuple quantification of fairness, privacy, and regulatory compliance for (a-b) only face-based datasets and (c-d) jointly with medical datasets. (a, c) The 3-D scatter plot of the different datasets across the three axes with the \textit{FPR dataset} plotted with perfect fairness, privacy preservation, and regulatory compliance. (b, d) The scatter plot after performing DBSCAN clustering with $eps=1$. We observe that the FB Fairness Dataset and the UTKFace dataset lie the closest to the \textit{FPR dataset}. }
\label{fig:clusteranalysis}
\end{figure*}

\begin{table*}[!t]
\caption{\label{tab:fprsummary}Summary of the different scores obtained for fairness, privacy, and regulatory compliance quantification obtained for the biometric datasets in the study. As can be clearly seen, the highest weighted average for fairness, privacy, and regulatory compliance is obtained on the FB Fairness dataset (Casual Conversations).}
\centering
\begin{tabular}{|p{0.4\textwidth}|C{0.11\textwidth}|C{0.11\textwidth}|C{0.11\textwidth}|C{0.11\textwidth}|}
\hline
\textbf{Dataset Name}                                                                    & \textbf{Fairness score (Max. 5)} & \textbf{Privacy Score (Max. 6)} & \textbf{Regulatory Score (Max. 3)} & \textbf{Weighted Average} \\ \hline
MORPH \cite{ricanek2006morph}                                                           & 2.2045                           & 3                               & 1                                  & 0.4247                    \\
Caltech 10K Web Faces \cite{caltech10kfaces}                                            & 0.0000                           & 6                               & 0                                  & 0.3333                    \\
FaceTracer \cite{kumar2008facetracer}                                                   & 0.6791                           & 4                               & 0                                  & 0.2675                    \\
CMU Multi-PIE \cite{ryan2009automated}                                                  & 0.2463                           & 5                               & 1                                  & 0.4053                    \\
PubFig \cite{kumar2009attribute}                                                        & 1.3349                           & 3                               & 0                                  & 0.2557                    \\
Plastic surgery database \cite{singh2010plastic}                                        & 0.0000                           & 5                               & 0                                  & 0.2778                    \\
Texas 3D Face recognition database \cite{gupta2010texas}                                & 1.1835                           & 5                               & 1                                  & 0.4678                    \\
ChokePoint \cite{wong2011patch}                                                         & 0.5097                           & 6                               & 1                                  & 0.4784                    \\
SCFace \cite{grgic2011scface}                                                           & 1.8257                           & 3                               & 0                                  & 0.2884                    \\
YouTube Faces \cite{wolf2011face}                                                       & 0.0000                           & 5                               & 0                                  & 0.2778                    \\
EGA \cite{riccio2012ega}                                                                & 0.7141                           & 5                               & 1                                  & 0.4365                    \\
10k US Adult Faces Database \cite{bainbridge2013intrinsic}                              & 0.5966                           & 5                               & 2                                  & 0.5398                    \\
Denver Intensity of Spontaneous Facial Action (DISFA) Database \cite{mavadati2013disfa} & 1.2085                           & 5                               & 1                                  & 0.4695                    \\
IMFDB-CVIT \cite{setty2013indian}                                                       & 1.3205                           & 3                               & 0                                  & 0.2547                    \\
SiblingsDB Database HQ Faces \cite{vieira2014detecting}                                 & 1.2643                           & 4                               & 1                                  & 0.4176                    \\
SiblingsDB Database LQ Faces \cite{vieira2014detecting}                                 & 0.6706                           & 5                               & 0                                  & 0.3225                    \\
Stirling 3D Face Dataset \cite{stirlingdb}                                              & 1.0040                           & 4                               & 1                                  & 0.4003                    \\
Adience \cite{eidinger2014age}                                                          & 0.5543                           & 5                               & 0                                  & 0.3147                    \\
CACD \cite{chen2014cross}                                                               & 1.2337                           & 3                               & 1                                  & 0.3600                    \\
CelebA \cite{liu2015deep}                                                               & 1.2262                           & 4                               & 0                                  & 0.3040                    \\
FaceScrub \cite{ng2014data}                                                             & 0.6670                           & 5                               & 0                                  & 0.3222                    \\
LFWA \cite{liu2015deep}                                                                & 1.3276                           & 3                               & 0                                  & 0.2552                    \\
MOBIO - Mobile Biometry Face and Speech Database \cite{tresadern2012mobile}             & 1.0501                           & 5                               & 1                                  & 0.4589                    \\
CAFE - The Child Affective Face Set \cite{lobue2015child}                               & 1.0261                           & 5                               & 2                                  & 0.5684                    \\
UFI - Unconstrained Facial Images \cite{lenc2015unconstrained}                          & 0.0000                           & 5                               & 0                                  & 0.2778                    \\
AFAD \cite{niu2016ordinal}                                                              & 1.1887                           & 5                               & 0                                  & 0.3570                    \\
IMDB-Wiki \cite{rothe2018deep}                                                          & 0.6797                           & 3                               & 1                                  & 0.3231                    \\
AgeDB \cite{moschoglou2017agedb}                                                        & 0.8603                           & 4                               & 0                                  & 0.2796                    \\
Large Age-Gap Database (LAG) \cite{bianco2017large}                                     & 0.3333                           & 4                               & 0                                  & 0.2444                    \\
PPB \cite{buolamwini2018gender}                                                         & 1.3274                           & 4                               & 0                                  & 0.3107                    \\
The IST-EURECOM Light Field Face Database \cite{sepas2017eurecom}                       & 1.2160                           & 4                               & 1                                  & 0.4144                    \\
UTK Face \cite{zhang2017age}                                                            & 2.7062                           & 5                               & 1                                  & 0.5693                    \\
VGGFace2 \cite{cao2018vggface2}                                                         & 0.0000                           & 4                               & 0                                  & 0.2222                    \\
Disguised Faces in the Wild (DFW) \cite{kushwaha2018disguised}                          & 0.0000                           & 6                               & 0                                  & 0.3333                    \\
IJB-C \cite{maze2018iarpa}                                                              & 2.1279                           & 4                               & 1                                  & 0.4752                    \\
IMDB-Face \cite{wang2018devil}                                                          & 0.4067                           & 4                               & 1                                  & 0.3604                    \\
LAOFIW \cite{alvi2018turning}                                                           & 0.0000                           & 5                               & 2                                  & 0.5000                    \\
RFW \cite{wang2019racial}                                                               & 1.4959                           & 5                               & 0                                  & 0.3775                    \\
VIP\_attribute Dataset \cite{dantcheva2018show}                                         & 0.6667                           & 5                               & 0                                  & 0.3222                    \\
AAF \cite{cheng2019exploiting}                                                          & 0.7449                           & 5                               & 0                                  & 0.3274                    \\
BUPT-Balancedface \cite{shi2020pv}                                                      & 0.3333                           & 5                               & 0                                  & 0.3000                    \\
BUPT-GlobalFace \cite{shi2020pv}                                                        & 0.3595                           & 5                               & 0                                  & 0.3017                    \\
DroneSURF \cite{kalra2019dronesurf}                                                     & 0.0000                           & 6                               & 1                                  & 0.4444                    \\
FairFace \cite{karkkainen2021fairface}                                                  & 2.5357                           & 5                               & 0                                  & 0.4468                    \\
Indian Institute of Science Indian Face Dataset (IISCIFD) \cite{katti2019you}           & 1.0570                           & 5                               & 2                                  & 0.5705                    \\
Injured Face v2 (IF-V2) database \cite{majumdar2019subclass}                            & 0.0000                           & 5                               & 0                                  & 0.2778                    \\
SOF dataset \cite{afifi2019afif4}                                                       & 1.3183                           & 5                               & 1                                  & 0.4768                    \\
BFW \cite{robinson2020face}                                                             & 0.6667                           & 5                               & 1                                  & 0.4333                    \\
DiveFace \cite{morales2020sensitivenets}                                                & 1.6636                           & 5                               & 0                                  & 0.3887                    \\
\textbf{FB Fairness dataset (Casual Conversations)} \cite{hazirbas2021towards}                   & 1.5561                           & 5                               & 3                                  & \textbf{0.7149}                    \\
MAAD-Face \cite{terhorst2021maad}                                                       & 0.9022                           & 4                               & 1                                  & 0.3935                    \\
Sejong Face Database (SFD) \cite{cheema2021sejong}                                      & 1.0566                           & 4                               & 1                                  & 0.4038                    \\ \hline
\end{tabular}
\end{table*}

\begin{table*}[]
\caption{\label{tab:fprsummarymedical}Summary of the different scores obtained for fairness, privacy, and regulatory compliance quantification obtained for the medical datasets in the study. The chest Xray datasets in this study provide a poor regulatory compliance score. The highest weighted average for fairness, privacy and regulatory compliance is obtained on the Montgomery County chest X-ray set (MC).}
\centering
\begin{tabular}{|p{0.34\textwidth}|C{0.13\textwidth}|C{0.13\textwidth}|C{0.13\textwidth}|C{0.13\textwidth}|}
\hline
\textbf{Dataset Name}                                                   & \textbf{Fairness score (Max. 5)} & \textbf{Privacy Score (Max. 6)} & \textbf{Regulatory Score (Max. 3)} & \textbf{Weighted Average} \\ \hline
\textbf{Montgomery County chest X-ray set (MC)} \cite{jaeger2014two}            & 1.4164                           & 4.00                            & 1.00                               & \textbf{0.4278}                    \\
Shenzhen chest X-ray set \cite{jaeger2014two}                          & 1.3300                           & 4.00                            & 1.00                               & 0.4220                    \\
ChestX-ray8 (NIH) \cite{wang2017chestx}                                & 1.1740                           & 4.00                            & 0.00                               & 0.3005                    \\
RSNA Pneumonia Challenge \cite{shih2019augmenting}                     & 1.1478                           & 5.00                            & 0.00                               & 0.3543                    \\
CheXpert (Stanford University) \cite{irvin2019chexpert}                & 1.5318                           & 4.00                            & 0.00                               & 0.3243                    \\
PadChest (University of Alicante) \cite{bustos2020padchest}            & 1.2094                           & 4.00                            & 1.00                               & 0.4140                    \\
BIMCV+ dataset \cite{vaya2020bimcv}                                    & 1.6556                           & 3.00                            & 1.00                               & 0.3882                    \\
COVID-19 Image Data Collection (CIDC) \cite{cohen2020covidProspective} & 1.2922                           & 3.00                            & 0.00                               & 0.2528                    \\ \hline
\end{tabular}
\end{table*}

\section{Results}
For this work, we surveyed a large number of datasets. Datasets containing human subjects were selected for the study. While fairness and privacy issues persist across different data domains such as objects and scenes \cite{rojasdollar,deng2009imagenet}, current regulatory norms are designed for people. While it is possible to extend the concepts presented in this study to other domains, we limit our discussion to face-based and medical imaging datasets.  After filtering through a total of 100 datasets and discarding datasets that are decommissioned, small in size (less than 100 images), and whose data could not be downloaded/accessed/requested, we were left with 60 datasets. These 60 datasets are used for the analysis and quantification of the responsible rubric. We use 52 face-based biometric datasets (Table \ref{tab:datasetdetails}), and eight chest Xray based medical datasets (Table \ref{tab:medicaldatasetdetails}). For face-based datasets, we filtered through over 120 datasets removing datasets that had been decommissioned, older than 2010, and whose data was inaccessible. For chest Xray datasets, we similarly surveyed through about 20 datasets before obtaining the eight analyzed in this work. We quantify the datasets across the dimensions of fairness, privacy, and regulatory compliance. Using the specified quantification methodology, we obtain a 3-tuple containing scores across the three dimensions. Analysis across the three different dimensions has been obtained through Fig.~\ref{fig:barplotsummary} where the distribution of scores has been plotted.\\

\noindent \textbf{Fairness in Datasets:} The fairness of datasets is calculated based on Eqn. \ref{eq:fairnessquant}. Representative and balanced datasets have been shown to provide fairer performance across different demographic subgroups \cite{wang2021meta}. The fairness metric described in this work provides a maximum value of 5, with five being the fairest. The average value for the fairness score obtained for the datasets comes out to be 0.96 $\pm$ 0.64, signifying that, on average, the fairness score of a dataset ranges from 0.32 to 1.6. The detailed results are provided in Tables ~\ref{tab:fprsummary} and ~\ref{tab:fprsummarymedical} for biometric and medical datasets, respectively. The UTKFace dataset is observed to be the fairest, with a score of 2.71 among the datasets listed here, providing maximum representation. It should be noted that with a maximum score of 5, the UTKFace dataset achieves slightly more than half that score. Interestingly, the average fairness score for the eight medical datasets was 1.34 $\pm$ 0.17 while the same score for biometric datasets came out to be 0.90 $\pm$ 0.67.\\

\noindent \textbf{Privacy Preservation in Datasets:} The privacy preserved in datasets is computed based on the presence of privacy-compromising information in the annotations, such as names of subjects and the presence of critical objects such as credit cards. A $P$ indicating the privacy preservation capacity and $PL$ indicating privacy leakage of the dataset are calculated. The distribution of $P$ for privacy quantification is presented in Fig. \ref{fig:barplotsummary}. The best value of $P$ is 6. We observe that the DroneSURF dataset does not contain any private information, which makes it perfectly privacy-preserving. The medical datasets in the study de-identify their subjects but naturally leak information about medical conditions, while some further provide sensitive information such as location.
\\

\noindent \textbf{Regulatory Compliance in Datasets: } With modern IT laws in place, the regulatory compliance of datasets is quantified based on institutional approval of the dataset, subject's consent to the data collection, facility for expungement/correction of the subject's data from the dataset. Based on these criteria, the compliance scores are calculated with a maximum value of three. The distribution of scores is provided in Fig. \ref{fig:barplotsummary}. On average, a regulatory score value of 0.58 is obtained. We observe that the FB Fairness Dataset (Casual Conversations) satisfies all regulatory compliances, thereby obtaining the maximum regulatory score, whereas most datasets provide a score of 0 or 1. \\

\noindent \textbf{Fairness-Privacy Paradox in Datasets:} Many face-based biometric datasets provide sensitive attribute information. This leads to a \textit{fairness-privacy paradox} where the presence of these annotations enables fairness quantification but leads to privacy leakage. One way to remedy the situation is by providing population statistics in the published dataset papers instead of sensitive attribute labels for each sample. However, current fairness algorithms are evaluated through sensitive attribute annotations in the dataset, and their absence can hinder the fairness evaluation process. In differential privacy-based solutions, it has been observed that the performance degradation is unequal across different subgroups~\cite{bagdasaryan2019differential}, highlighting the need for labels for fairness analysis. The \textit{fairness-privacy paradox} remains an open problem for datasets containing sensitive attribute information such as biometrics and medical imaging. With ongoing discussion regarding concerns for privacy and fairness, regulations can sometimes provide conflicting guidance on privacy laws and proposed AI laws, giving researchers and industry a reason to approach this paradox with caution in dataset development. Recent work in face recognition is exploring models trained using synthetically generated datasets\cite{qiu2021synface, melzi2023gandiffface, kim2023dcface}. However, the training of powerful generative models utilizes large face datasets. Some diffusion-based models have also been shown to replicate the training data during generation.

\noindent \textbf{Holistic View of Responsibility in Datasets:} When the aforementioned factors are studied in conjunction, we obtain a three-dimensional representation of the datasets. The 3-tuple provides insight into how responsible a dataset may be considered for downstream training. To observe the behavior of the 3-tuple visually, we plotted a 3-D scatter plot for the face datasets along with a hypothetical \textit{FPR dataset} (Fig. \ref{fig:clusteranalysis}(a)). The hypothetical FPR dataset has a perfect fairness, privacy, and regulatory score. After applying the DBSCAN algorithm with $eps=1$ (the maximum distance between two points to be considered as a part of one cluster), we observe five clusters with two outliers. The FB Fairness Dataset and the UTKFace dataset come out to be outliers with a Euclidean distance of 3.59 and 3.20 units from the \textit{FPR dataset}. 
When compared to the other clusters, we observe that FB Fairness Dataset and the UTKFace dataset lie the closest to the \textit{FPR dataset}.

Other cluster centers lie at a distance of 4.56, 4.79, 5.11, 5.20, and 5.33 units from the \textit{FPR dataset}, with the clusters containing 4, 7, 3, 32, and 4 points, respectively. The next closest cluster is formed by the LAOFIW, 10k US Adult Faces Database, CAFE Database, and IISCIFD datasets with average scores of 0.67, 5, and 2 for fairness, privacy, and regulatory compliance, respectively. Similar observations can be made when the scatter plot includes medical datasets along with the face datasets(Fig. \ref{fig:clusteranalysis}(c-d)). The  numerical results are tabulated in Tables \ref{tab:fprsummary} and \ref{tab:fprsummarymedical}. A weighted average of the three scores is calculated by dividing each score by its maximum value and then taking an average that provides a value in the range of 0 to 1 (Table \ref{tab:fprsummary}). By utilizing this average, we observe that the top three responsible datasets come out to be the FB Fairness dataset (Casual Conversations), the Indian Institute of Science Indian Face Dataset (IISCIFD), and the UTKFace dataset. A high regulatory compliance score plays an important role in the overall responsibility score of FB Fairness and IISCIFD datasets. In contrast, a high fairness score imparts UTKFace a high responsible rubric value. To summarize the observations made over the existing face datasets, we find that-
\begin{itemize}
    \item Most of the existing datasets suffer on all the three axes of \textit{fairness}, \textit{privacy} and \textit{regulatory compliance} as per the proposed metric. For example, the UTKFace dataset is among the fairest datasets but performs poorly on regulatory compliance. On the other hand, the LFWA dataset lacks on all three fronts- fairness, privacy preservation, and regulatory compliance.
    \item While many works claim fairness as the primary focus in their datasets, these datasets provide poor fairness scores on evaluation. One such example is the DiveFace dataset. The fairness quantification of datasets using our framework shows that being fair is a major concern with 91\% of the existing datasets obtaining a fairness score of two or less out of five.
    \item A vast number of large-scale  datasets in Computer Vision are web-curated without any institutional approval. These datasets are often released under various CC-BY licenses even when these datasets do not have subject consent. We found that these datasets also fares low on the fairness front since the annotations are not always reliable, posing major risks to overall data responsibility.
    \item Following regulatory norms effectively improves the responsibility rubric for a given dataset; however, most datasets are not compliant based on the available information with 89\% datasets having a compliance score of 0 or 1.
    \item When comparing fairness, privacy, and regulatory scores, it is clear that the privacy scores are higher in general. It is worth noting that privacy standards and constraints are already defined and have  existed for a few years now~\cite{regulation2016regulation}, and datasets are possibly collected with these regulations in mind. This further indicates a need for fairness and regulatory constraints that promote data collection with higher fairness and regulatory standards.  \\
\end{itemize}


\noindent \textbf{Recommendations:} Based on the observations of our framework on a large number of datasets, we provide certain recommendations to aid better dataset collection in the future.
\begin{itemize}
    \item \textit{Institutional Approval, Ethics Statement, and Subject Consent:} Datasets involving human subjects should receive approval from the institutional review board (such as IRB in the United States). Future regulations may require consent from subjects to be obtained explicitly for the dataset and its intended use. 
    \item \textit{Facility for Expungement/Correction of Subject's Data:} Datasets should provide a facility to contact the dataset owners to remove and/or correct information concerning the subject. This is necessary to be compliant with data privacy laws such as GDPR. Some existing datasets already provide the facility for expungement in their datasets such as the FB Fairness Dataset, IJB-C and the UTKFace datasets. 
    \item \textit{Fairness and Privacy:} Datasets should be collected from a diverse population, and distribution across sensitive attributes should be provided while being privacy-preserving. The proposed fairness and privacy scores can aid in quantifying a dataset's diversity and privacy preservation.  
    \item \textit{Datasheet:} Datasets should curate and provide a datasheet containing information regarding the objectives, intended use, funding agency, the demographic distribution of subjects/images, licensing information, and limitations of the dataset. By specifying intended use, the data can be restricted for processing outside of intended use under the GDPR. An excellent resource for the construction of datasheets is provided by Gebru et al.~\cite{gebru2021datasheets}. We propose modifications in the datasheet by Gebru et al. by adding questions concerning fairness, privacy, and regulatory compliance in datasets (Refer Tables \ref{tab:datasheet} and \ref{tab:newdatasheet}).
\end{itemize}

\noindent \textbf{Limitations:} The formulation for quantification in this work considers a dataset fair based on the distribution of its labels.  However, we do not account for the diversity of the data such as the presence of duplicate images for particular subgroups. Further, we do not comment on equity vs equality in the distribution of images. We note that it may be desirable to have unequal distribution between groups (e.g., when one group is harder to process than others and requires more data for the model to reach equal performance across groups) for some applications. Further, the current formulation for fairness, privacy, and regulatory scores is provided for datasets constituting individuals. While object-based datasets may also suffer from fairness issues, current data regulations are designed in accordance with the impact on human individuals. We leave analysis on object-based datasets for future work. Finally, we would like to note that the recommendations and datasheets introduced in this work are intended to establish the highest standards which can be challenging to achieve given the capabilities of current technologies. These recommendations are meant to serve as a ``north star" and reaching them requires deliberate research effort. The fairness-privacy paradox remains an open problem in the community. Similarly, removing instances of data from already trained models requires \textit{unlearning} techniques, which while being actively explored, are far from being perfect.

\section{Conclusion}

While the vast majority of the existing literature focuses on the design of trustworthy machine learning algorithms, in this work, we offer a fresh perspective for evaluating reliability through a discussion of responsible datasets with respect to fairness, privacy, and regulatory compliance. A detailed analysis is performed  for face-based and chest Xray image datasets. We further provide recommendations for the design of `responsible ML datasets.' With governments around the world regularizing data protection laws, the method for the creation of datasets in the scientific community requires revision, and it is our assertion that the proposed quantitative measures, qualitative datasheets, and recommendations can stimulate the creation of responsible datasets which can lead to building responsible AI systems.

\section{Acknowledgements}
This preprint has not undergone any post-submission improvements or corrections. The Version of Record of this article is published in Nature Machine Intelligence, and is available online at \url{https://doi.org/10.1038/s42256-024-00874-y}.

\begin{table*}[]
\caption{\label{tab:datasheet} Research questions to account for the preparation of datasets. Parts of this table are taken from Gebru et al.\cite{gebru2021datasheets}. The proposed questions are highlighted in \textbf{bold}.}
\centering
\begin{tabular}{|p{0.1\textwidth}|p{0.85\textwidth}|}
\hline
\textbf{Category}                                & \textbf{Research Questions}                                                                                                                                                                                               \\ \hline
\multirow{3}{*}{Motivation}                      & For what purpose was the dataset created?                                                                                                                                                                                 \\
                                                 & \textbf{Who created the dataset (e.g., which team, research group) and on behalf of which entity (e.g., company, institution, organization)?}                                                                             \\
                                                 & \textbf{Who funded the creation of the dataset?}                                                                                                                                                                          \\ \hline
\multirow{13}{*}{Composition}                    & What do the instances that comprise the dataset represent (e.g., documents, photos, people, countries)?                                                                                                                   \\
                                                 & \textbf{How many instances are there in total (of each type, if appropriate)?}                                                                                                                                            \\
                                                 & \textbf{What is the distribution of the instances with respect to the target or label?}                                                                                                                                   \\
                                                 & \textbf{Is the distribution of instances in the dataset long-tailed  with respect to the target or label?}                                                                                                                \\
                                                 & Does the dataset contain all possible instances or is it a sample (not necessarily random) of instances from a larger set?                                                                                                \\
                                                 & What data does each instance consist of?                                                                                                                                                                                  \\
                                                 & Is there a label or target associated with each instance?                                                                                                                                                                 \\
                                                 & \textbf{Is any information missing from individual instances?}                                                                                                                                                            \\
                                                 & Are relationships between individual instances made explicit (e.g., users’ movie ratings, social network links)?                                                                                                          \\
                                                 & Are there recommended data splits (e.g., training, development/validation, testing)?                                                                                                                                      \\
                                                 & Is the dataset self-contained, or does it link to or otherwise rely on external resources (e.g., websites, tweets, other datasets)?                                                                                       \\
                                                 & \textbf{If biometric (or sensitive) information is present in the instances of the dataset such as fingerprint or faces, are these annotated?}                                                                            \\
                                                 & \textbf{If the biometric (or sensitive) information is not annotated, is it discussed in detail in the accompanying research paper or elsewhere?}                                                                         \\ \hline
\multirow{5}{2cm}{Collection Process}              & If the dataset is a sample from a larger set, what was the sampling strategy (e.g., deterministic, probabilistic with specific sampling probabilities)?                                                                   \\
                                                 & \textbf{How was the data associated with each instance acquired?}                                                                                                                                                         \\
                                                 & Who was involved in the data collection process (e.g., students, crowdworkers, contractors) and how were they compensated (e.g., how much were crowdworkers paid)?                                                        \\
                                                 & \textbf{What is the demographic of the people involved in the collection and annotation process?}                                                                                                                         \\
                                                 & Over what timeframe was the data collected?                                                                                                                                                                               \\ \hline
\multirow{3}{2cm}{Preprocessing/ cleaning/ labeling} & \textbf{Was any preprocessing/cleaning/labeling of the data done (e.g., discretization or bucketing, tokenization, part-of-speech tagging, SIFT feature extraction, removal of instances, processing of missing values)?} \\
                                                 & \textbf{Was the “raw” data saved in addition to the preprocessed/cleaned/labeled data (e.g., to support unanticipated future uses)?}                                                                                      \\
                                                 & Is the software that was used to preprocess/clean/label the data available?                                                                                                                                               \\ \hline
\multirow{6}{*}{Distribution}                    & \textbf{Will the dataset be distributed to third parties outside of the entity (e.g., company, institution, organization) on behalf of which the dataset was created?}                                                    \\
                                                 & How will the dataset will be distributed (e.g., tarball on website, API, GitHub)?                                                                                                                                         \\
                                                 & When will the dataset be distributed?                                                                                                                                                                                     \\
                                                 & \textbf{Will the dataset be distributed under a copyright or other intellectual property (IP) license, and/or under applicable terms of use (ToU)?}                                                                       \\
                                                 & Have any third parties imposed IP-based or other restrictions on the data associated with the instances?                                                                                                                  \\
                                                 & Is there a repository that links to any or all papers or systems that use the dataset?                                                                                                                                    \\ \hline
\multirow{8}{*}{Maintenance}                     & Who will be supporting/hosting/maintaining the dataset?                                                                                                                                                                   \\
  & How can the owner/curator/manager of the dataset be contacted (e.g., email address)?                                                                                                                                      \\
  & \textbf{Who bears responsibility in case of a violation of rights?}                                                                                                                                                       \\
                                                 & Is there an erratum?                                                                                                                                                                                                      \\
                                                 & Will the dataset be updated (e.g., to correct labeling errors, add new instances, delete instances)?                                                                                                                      \\
                                                 & Will older versions of the dataset continue to be supported/hosted/maintained?                                                                                                                                            \\
                                                 & If others want to extend/augment/build on/contribute to the dataset, is there a mechanism for them to do so?                                                                                                              \\
                                                 & \textbf{Are there specific guidelines for derivative work such as allowed/disallowed derivatives, distribution and licensing requirements?}                                                                               \\ \hline
\end{tabular}
\end{table*}


\begin{table*}[]
\caption{\label{tab:newdatasheet} Research questions surrounding the fairness, privacy and regulatory compliance of datasets. Parts of this table are taken from Gebru et al.\cite{gebru2021datasheets}. The proposed questions are highlighted in \textbf{bold}.}
\centering
\begin{tabular}{|p{0.1\textwidth}|p{0.85\textwidth}|}
\hline
\multicolumn{1}{|l|}{\textbf{}} & \textbf{Research Questions}                                                                                                  \\ \hline
\multirow{11}{*}{Fairness}       & Does the dataset identify any subpopulations (e.g., by age, gender)?  \\
& \textbf{Which subpopulations are identified in the dataset and how many subgroups of each subpopulation are represented? (inclusivity score)}                                                \\
 & \textbf{What is the distribution of the subgroups for each of the subpopulations? (diversity score)}       \\
& What mechanisms or procedures were used to collect the data (e.g., hardware apparatuses or sensors, manual human curation, software programs, software APIs)? \\
& \textbf{How reliable is the source of annotation for subpoulations and were the labels self-reported, apparent or classifier-generated? (label score)} \\
& \textbf{Based on the annotation in the dataset for the sensitive attributes, what is the inclusivity score, diversity score, label score and the overall fairness score?}  \\
 & \textbf{Are there any correlations between the sensitive attributes or with other attributes? (shortcuts/spurious correlations)} \\ \hline
\multirow{5}{*}{Privacy}        & Does the dataset contain data that might be considered confidential (e.g., data that is protected by legal privilege or by doctor-patient confidentiality, data that includes the content of individuals’ non-public communications)?                                                                                                                                    \\
                                & Does the dataset contain data that, if viewed directly, might be offensive, insulting, threatening, or might otherwise cause anxiety?                                                                                                                                                                                                                                    \\
& Is it possible to identify individuals (i.e., one or more natural persons), either directly or indirectly (i.e., in combination with other data) from the dataset?\\
& \textbf{If the data has been de-identified, what mechanism has been used for de-identification?} \\
 & Does the dataset contain data that might be considered sensitive in any way (e.g., data that reveals race or ethnic origins, sexual orientations, religious beliefs, political opinions or union memberships, or locations; financial or health data; biometric or genetic data; forms of government identification, such as social security numbers; criminal history)? \\ \hline
\multirow{13}{*}{Regulatory}    & Were any ethical review processes conducted (e.g., by an institutional review board)? \\
& Did you collect the data from the individuals in question directly, or obtain it via third parties or other sources (e.g., websites)?\\
& Were the individuals in question notified about the data collection?  \\
& Did the individuals in question consent to the collection and use of their data? \\
 & If consent was obtained, were the consenting individuals provided with a mechanism to revoke their consent in the future or for certain uses?  \\
 & \textbf{Irrespective of consent, is there a mechanism to expunge the data of individuals on request?}\\
 & What tasks has the dataset been used for (if any) and what other tasks can the dataset be used for? \\
& Are there tasks for which the dataset should not be used?  \\
 & If the dataset relates to people, are there applicable limits on the retention of the data associated with the instances (e.g., were the individuals in question told that their data would be retained for a fixed period of time and then deleted)? \\
& \textbf{In case of an expungement request of data, what happens to the model trained using the previous data (if any)?} \\
& \textbf{Is there a mechanism to extract the data from the trained models?} \\
& Do any export controls or other regulatory restrictions apply to the dataset or to individual instances?\\
 & \textbf{Does the dataset follow the applicable data privacy laws in your region such as GDPR in EU (if any)?}  \\ \hline
\multirow{3}{*}{Future Impact}  & Has an analysis of the potential impact of the dataset and its use on data subjects (e.g., a data protection impact analysis) been conducted?                                                                                                                                                                                                                            \\
                                & Is there anything about the composition of the dataset or the way it was collected and preprocessed/cleaned/labeled that might impact future uses?                                                                                                                                                                                                                       \\
                                & \textbf{Could there be any ethical concerns raised for the data?}                                                                                                                                                                                                                                                                                                        \\ \hline
\end{tabular}
\end{table*}

\begin{table*}[!t]
\caption{\label{tab:datasetdetails}Details of the biometric datasets employed for the study. The \textit{Demographic Attributes} specify the annotations for attributes such as ethnicity, skin tone, age, and gender.}
\centering
\scriptsize
\begin{tabular}{|p{0.29\textwidth}|C{0.025\textwidth}|C{0.09\textwidth}|p{0.19\textwidth}|p{0.2\textwidth}|C{0.06\textwidth}|}
\hline
\textbf{Dataset Name}                                                                    & \textbf{Year} & \textbf{Images}             & \textbf{Facial Tasks}                                                          & \textbf{Demographic Attributes}                     & \textbf{Web Collected} \\ \hline
MORPH \cite{ricanek2006morph}                                                           & 2006          & 400K                        & Recognition                                                                    & Ethnicity, Gender, Age                              & No                     \\
Caltech 10K Web Faces \cite{caltech10kfaces}                                            & 2007          & 10.5K                       & Detection                                                                      & -                                                   & Yes                    \\
FaceTracer \cite{kumar2008facetracer}                                                   & 2008          & 3.1M                        & Attribute Classification                                                       & Ethnicity, Gender, Age                              & Yes                    \\
CMU Multi-PIE \cite{ryan2009automated}                                                  & 2009          & 755K                        & Recognition                                                                    & Ethnicity, Gender                                   & No                     \\
PubFig \cite{kumar2009attribute}                                                        & 2009          & 58.7K                       & Verification                                                                   & Ethnicity, Gender, Age                              & Yes                    \\
Plastic surgery database \cite{singh2010plastic}                                        & 2010          & 900 image pairs             & Verification                                                                   & -                                                   & Yes                    \\
Texas 3D Face recognition database \cite{gupta2010texas}                                & 2010          & 1.1K                        & Recognition                                                                    & Ethnicity, Gender, Age                              & No                     \\
ChokePoint \cite{wong2011patch}                                                         & 2011          & 64.2 K images, 48 videos    & Person Recognition                                                             & Gender                                              & No                     \\
SCFace \cite{grgic2011scface}                                                           & 2011          & 4.1K                        & Recognition                                                                    & Ethnicity (Caucasian), Gender, Age                  & No                     \\
YouTube Faces \cite{wolf2011face}                                                       & 2011          & 3.4K videos                 & Recognition                                                                    & -                                                   & Yes                    \\
EGA \cite{riccio2012ega}                                                                & 2012          & 2.3K                        & Recognition                                                                    & Ethnicity, Gender, Age                              & No                     \\
MOBIO - Mobile Biometry Face and Speech Database \cite{tresadern2012mobile}             & 2012          & 75 videos                   & Recognition                                                                    & Gender                                              & No                     \\
10k US Adult Faces Database \cite{bainbridge2013intrinsic}                              & 2013          & 2.2K                        & Recognition                                                                    & Ethnicity, Gender                                   & Yes                    \\
Denver Intensity of Spontaneous Facial Action (DISFA) Database \cite{mavadati2013disfa} & 2013          & 130K                        & Expression Recognition                                                         & Ethnicity, Gender                                   & No                     \\
IMFDB-CVIT \cite{setty2013indian}                                                       & 2013          & 34.5K                       & Recognition                                                                    & Ethnicity (Indian), Gender, Age                     & Yes                    \\
Stirling 3D Face Dataset \cite{stirlingdb}                                              & 2013          & 2K (subset)                 & 3D Reconstruction                                                              & Gender                                              & No                     \\
Adience \cite{eidinger2014age}                                                          & 2014          & 26.5K                       & Age and Gender Classification                                                  & Gender, Age                                         & Yes                    \\
CACD \cite{chen2014cross}                                                               & 2014          & 160K                        & Recognition                                                                    & Age                                                 & Yes                    \\
FaceScrub \cite{ng2014data}                                                             & 2014          & 106.8K                      & Recognition                                                                    & Gender                                              & Yes                    \\
SiblingsDB Database HQ Faces \cite{vieira2014detecting}                                 & 2014          & 0.4K                        & Verification                                                                   & Ethnicity (Caucasian), Gender, Age                  & No                     \\
SiblingsDB Database LQ Faces \cite{vieira2014detecting}                                 & 2014          & 0.2K                        & Verification                                                                   & Ethnicity (Caucasian), Gender                       & Yes                    \\
CAFE - The Child Affective Face Set \cite{lobue2015child}                               & 2015          & 1.2K                        & Expression Recognition                                                         & Ethnicity, Gender, Age (Children)                   & No                     \\
CelebA \cite{liu2015deep}                                                               & 2015          & 200K                        & Detection, Landmark Detection, Editing and Synthesis                           & Skin-tone, Gender, Age                              & Yes                    \\
LFWA \cite{liu2015deep} (40 attributes)                                                              & 2015          & 200K                        & Detection, Attribute Classification, Landmark Detection, Editing and Synthesis & Skin-tone, Gender, Age                              & Yes                    \\
UFI - Unconstrained Facial Images \cite{lenc2015unconstrained}                          & 2015          & 4.3K                        & Identification                                                                 & -                                                   & Yes                    \\
AFAD \cite{niu2016ordinal}                                                              & 2016          & 160K                        & Age Estimation                                                                 & Ethnicity (East Asian), Gender, Age                 & Yes                    \\
AgeDB \cite{moschoglou2017agedb}                                                        & 2017          & 16.4K                       & Recognition, Age Estimation                                                    & Gender, Age                                         & Yes                    \\
Large Age-Gap Database (LAG) \cite{bianco2017large}                                     & 2017          & 3.8K                        & Verification                                                                   & Age                                                 & Yes                    \\
The IST-EURECOM Light Field Face Database \cite{sepas2017eurecom}                       & 2017          & 4K                          & Recognition                                                                    & Gender, Age                                         & No                     \\
UTK Face \cite{zhang2017age}                                                            & 2017          & 20K                         & Detection, Age Estimation and Progression, Landmark Detection                  & Ethnicity, Gender, Age                              & Yes                    \\
Disguised Faces in the Wild (DFW) \cite{kushwaha2018disguised}                          & 2018          & 11.1K                       & Recognition                                                                    & Ethnicity (Indian, Caucasian)                       & Yes                    \\
IJB-C \cite{maze2018iarpa}                                                              & 2018          & 31.3 k images, 11.7K videos & Detection, Recognition, Verification, Clustering                               & Skin-tone, Gender, Age                              & Yes                    \\
IMDB-Face \cite{wang2018devil}                                                          & 2018          & 1.7M                        & Recognition                                                                    & Gender, Age                                         & Yes                    \\
IMDB-Wiki \cite{rothe2018deep}                                                          & 2018          & 523K                        & Age Estimation                                                                 & Gender, Age                                         & Yes                    \\
LAOFIW \cite{alvi2018turning}                                                           & 2018          & 14K                         & Attribute Classification                                                       & Ethnicity                                           & Yes                    \\
PPB \cite{buolamwini2018gender}                                                         & 2018          & 1.3K                        & Attribute Classification                                                       & Skin-tone, Gender                                   & Yes                    \\
VGGFace2 \cite{cao2018vggface2}                                                         & 2018          & 3.3M                        & Recognition                                                                    & -                                                   & Yes                    \\
VIP\_attribute Dataset \cite{dantcheva2018show}                                         & 2018          & 1K                          & Attribute Classification                                                       & Gender                                              & Yes                    \\
AAF \cite{cheng2019exploiting}                                                          & 2019          & 13.2K                       & Age Estimation, Gender Classification.                                         & Ethnicity (Asian), Gender, Age                      & Yes                    \\
DroneSURF \cite{kalra2019dronesurf}                                                     & 2019          & 411K images, 200 videos     & Detection, Recognition                                                         & Age                                                 & No                     \\
Indian Institute of Science Indian Face Dataset (IISCIFD) \cite{katti2019you}           & 2019          & 1.6K                        & Attribute Classification                                                       & Ethnicity (North Indian, South Indian), Gender, Age & Yes                    \\
Injured Face v2 (IF-V2) database \cite{majumdar2019subclass}                            & 2019          & 0.9K                        & Identification                                                                 & -                                                   & Yes                    \\
RFW \cite{wang2019racial}                                                               & 2019          & 40K                         & Recognition, Landmark Detection                                                & Ethnicity, Gender                                   & Yes                    \\
SOF dataset \cite{afifi2019afif4}                                                       & 2019          & 42.5K                       & Attribute Classification, Detection, Recognition                               & Gender, Age                                         & No                     \\
BFW \cite{robinson2020face}                                                             & 2020          & 20K                         & Recognition                                                                    & Ethnicity, Gender                                   & Yes                    \\
BUPT-Balancedface \cite{shi2020pv}                                                      & 2020          & 1.3M                        & Recognition                                                                    & Ethnicity                                           & Yes                    \\
BUPT-GlobalFace \cite{shi2020pv}                                                        & 2020          & 2M                          & Recognition                                                                    & Ethnicity                                           & Yes                    \\
DiveFace \cite{morales2020sensitivenets}                                                & 2020          & 120K                        & Recognition                                                                    & Ethnicity, Gender                                   & Yes                    \\
FairFace \cite{karkkainen2021fairface}                                                  & 2021          & 108.5K                      & Attribute Classification                                                       & Ethnicity, Gender, Age                              & Yes                    \\
FB Fairness dataset (Casual Conversations) \cite{hazirbas2021towards}                   & 2021          & 45.1K videos                & Attribute Classification, Face Manipulation                                    & Skin-tone, Gender, Age                              & No                     \\
MAAD-Face \cite{terhorst2021maad}                                                       & 2021          & 3.3M                        & Attribute Classification                                                       & Ethnicity, Gender, Age                              & Yes                    \\
Sejong Face Database (SFD) \cite{cheema2021sejong}                                      & 2021          & 24.5K                       & Recognition                                                                    & Ethnicity, Gender                                   & No                     \\ \hline
\end{tabular}
\end{table*}

\begin{table*}[!t]
\caption{\label{tab:medicaldatasetdetails}Table containing details of the medical datasets employed for the study. The \textit{Demographic Attributes} specify the annotations for attributes such as ethnicity, skin tone, age, and gender.}
\centering
\scriptsize
\begin{tabular}{|p{0.28\textwidth}|C{0.025\textwidth}|C{0.06\textwidth}|p{0.18\textwidth}|C{0.15\textwidth}|C{0.09\textwidth}|}\hline
\textbf{Dataset Name}                                                   & \multicolumn{1}{c|}{\textbf{Year}} & \textbf{Images} & \textbf{Tasks}                  & \textbf{Demographic Attributes} & \textbf{Web Collected} \\ \hline
Montgomery County chest X-ray set (MC) \cite{jaeger2014two}            & 2014                               & 0.01K           & Tuberculosis Detection          & Gender, Age                     & No                     \\ 
Shenzhen chest X-ray set \cite{jaeger2014two}                          & 2014                               & 0.7K            & Tuberculosis Detection          & Gender, Age                     & No                     \\ 
ChestX-ray8 (NIH) \cite{wang2017chestx}                                & 2017                               & 108K            & Thoracic Disease Classification & Gender, Age                     & No                     \\ 
RSNA Pneumonia Challenge \cite{shih2019augmenting}                     & 2018                               & 30K             & Pneumonia Detection             & Gender, Age                     & No                     \\ 
CheXpert (Stanford University) \cite{irvin2019chexpert}                & 2019                               & 224K            & Disease Classification          & Gender, Age                     & No                     \\ 
PadChest (University of Alicante) \cite{bustos2020padchest}            & 2019                               & 160K            & Disease Classification          & Gender, Age                     & No                     \\ 
BIMCV+ dataset \cite{vaya2020bimcv}                                    & 2020                               & 2.4K            & Covid-19 Detection              & Gender, Age                     & No                     \\ 
COVID-19 Image Data Collection (CIDC) \cite{cohen2020covidProspective} & 2020                               & 0.95K           & Covid-19 Detection              & Gender, Age                     & Yes                    \\ 
\hline
\end{tabular}
\end{table*}


\end{document}